\documentclass[authoryear,preprint,review,12pt]{elsarticle}

\usepackage{svg}
\usepackage{amssymb}
\usepackage{graphicx}
\usepackage{float}
\usepackage{amsthm}
\usepackage{mathtools}
\usepackage[toc,page, title, titletoc]{appendix}
\usepackage{booktabs}
\usepackage[hidelinks]{hyperref} 
\usepackage{subfig}
\usepackage[section]{placeins}
\usepackage{lineno}
\journal{Remote Sensing of Environment}
\begin{document}
\begin{frontmatter}

 \title{Scalable Geospatial Machine Learning for Power-Line Asset Risk: Integrating Remote Sensing for Lightning and Vegetation Risk Modelling}

\author[SAPN2]{Artur Sokolovsky}

\affiliation[SAPN2]{organization={SA Power Networks},%Department and Organization
            addressline={1 Anzac Highway, Keswick SA}, 
            city={Adelaide},
            postcode={5035}, 
            country={Australia}}

% \affiliation[SAPN]{organization={SA Power Networks},%Department and Organization
%             addressline={1 Richmond Road, Keswick SA}, 
%             city={Adelaide},
%             postcode={5035}, 
%             country={Australia}}

\author[SAPN2]{Bhavik Merai}
\author[SAPN2]{Moe Jafari}
\author[SAPN2]{Muen Chen}

\begin{abstract}
Electric power networks are increasingly exposed to weather-sensitive failure mechanisms that require asset-level, spatially explicit risk modelling for effective intervention planning. This study contributes a modular, robust, and explainable probability-of-failure (PoF) modelling framework for utility asset management. The central contribution is an asset-level architecture that can be scaled to new environmental data sources and additional PoF types without reworking the underlying pipeline. This is particularly relevant for industry settings, where risk models must remain operationally maintainable while adapting to changing data availability, asset-management priorities, and climate-driven hazard conditions.

We demonstrate the framework for vegetation-related and lightning-related failure modes using a harmonised geospatial machine-learning pipeline. The implementation integrates multi-source predictors, including topography (SRTM), vegetation condition (MODIS Normalised Difference Vegetation Index - NDVI), lightning climatology (LIS VHRMC), OpenStreetMap-derived proximity features, and utility operational records. The resulting architecture is computationally efficient, operationally extensible, and suitable for utility-scale deployment. It provides actionable asset-level risk stratification for inspection prioritisation, vegetation management, asset hardening, and resilience planning, supporting earlier intervention and more climate-resilient network operations.
\end{abstract}


\end{frontmatter}

%\linenumbers

\section{Introduction}

\subsection{Background}
Electric power networks are increasingly exposed to weather-sensitive failure mechanisms that are spatially heterogeneous and operationally costly. In this study, we focus on two hazard-specific probabilities of failure (PoF): \emph{lightning PoF}, defined as the probability that an asset fails due to lightning-related exposure, and \emph{vegetation PoF}, defined as the probability that an asset fails due to vegetation-related interactions (e.g., mechanical impact, or flashover events). Prior evidence shows that severe-weather impacts are non-uniform across networks, motivating asset-level modelling rather than aggregate system averages \citep{Mukherjee2018WeatherOutages}. 

From an environmental-process perspective, both hazards are dynamic. Lightning regimes may shift under warming climatic conditions \citep{Romps2014LightningClimate}, while vegetation dryness is also influenced by climate variability and change \citep{Abatzoglou2016WildfireClimate}. Geospatial Earth-observation products therefore provide an important foundation for scalable risk modelling: terrain information from SRTM \citep{Farr2007SRTM}, vegetation-condition signals from MODIS indices \citep{Didan2021MOD13Q1C61}, and long-term lightning climatology from LIS \citep{Cecil2014LISClimatology}. NDVI and related indices are particularly relevant as interpretable proxies of vegetation status in disturbance-oriented applications \citep{Pettorelli2005NDVI}.

\subsection{The Research Gap}
Despite progress in outage and hazard modelling, three limitations persist in the literature and in many operational implementations. First, studies frequently treat lightning and vegetation risks separately, limiting integrated asset-prioritisation decisions when both mechanisms can affect the same infrastructure. Second, many models remain too coarse spatially to support asset-level interventions (e.g., targeted inspection, trimming, hardening, or surge-protection planning). Third, hazard proxies are often not tightly coupled with utility failure records, reducing practical predictive value and limiting explainability for field deployment \citep{Mukherjee2018WeatherOutages}. 

Accordingly, there remains a need for a scalable and extensible PoF modelling design that uses harmonised geospatial predictors and observed utility failure outcomes, while supporting hazard-specific model instances for vegetation, lightning, and additional failure modes (e.g., animal-related faults).

\subsection{Research Question}
This study asks whether a scalable and extensible asset-level PoF framework, built on harmonised geospatial predictors and utility failure records, can produce operationally useful risk ranking and interpretable hazard-specific outputs for utility decision support. Under this question, the primary objective is methodological and operational validation within the available network context, rather than strict forward-in-time forecasting or geographic-domain transfer testing.

\subsection{Contributions}
This study makes four main contributions to asset-risk modelling for utility networks.
\begin{itemize}
\item \textbf{Modular multi-PoF formulation:} We develop an asset-level framework with a shared geospatial--machine-learning pipeline and hazard-specific model instances, demonstrated for vegetation PoF and lightning PoF, enabling consistent cross-hazard risk assessment without forcing a single combined model.
\item \textbf{Efficient and scalable modelling pipeline:} We implement a computationally efficient approach suitable for utility-scale deployment, combining multi-scale spatial feature engineering with a modelling architecture that can be operationalised across asset fleets of millions of assets.
\item \textbf{Extensible data and target design:} The framework is designed to be modular, allowing utilities to introduce new target variables (e.g., animal-related PoF), incorporate additional data sources, and expand the feature space as new operational and environmental datasets become available.
\item \textbf{Operationally meaningful asset-level granularity:} Results indicate that asset-level PoF modelling captures historical failure patterns with useful fidelity, supporting more targeted and effective interventions for maintenance prioritisation and resilience planning.
\end{itemize}

\subsection{Outline of the Study}
The remainder of this paper is organised as follows. Section~\ref{sec:methods} describes the data sources, preprocessing pipeline, spatial feature engineering, and machine-learning architecture for modular multi-PoF estimation using hazard-specific models. Section~\ref{sec:results} reports model validation, spatial risk patterns, explainability analyses, and operational risk stratification outputs. Section~\ref{sec:Discussion} interprets the findings, discusses limitations and practical implications, and outlines future directions toward near-real-time risk modelling. A glossary of acronyms is provided in Appendix~\ref{app:glossary}.

\section{Materials and Methods}\label{sec:methods}

\subsection{Data sources}

\begin{figure}[H]
\centering
\includegraphics[width=1.0\textwidth]{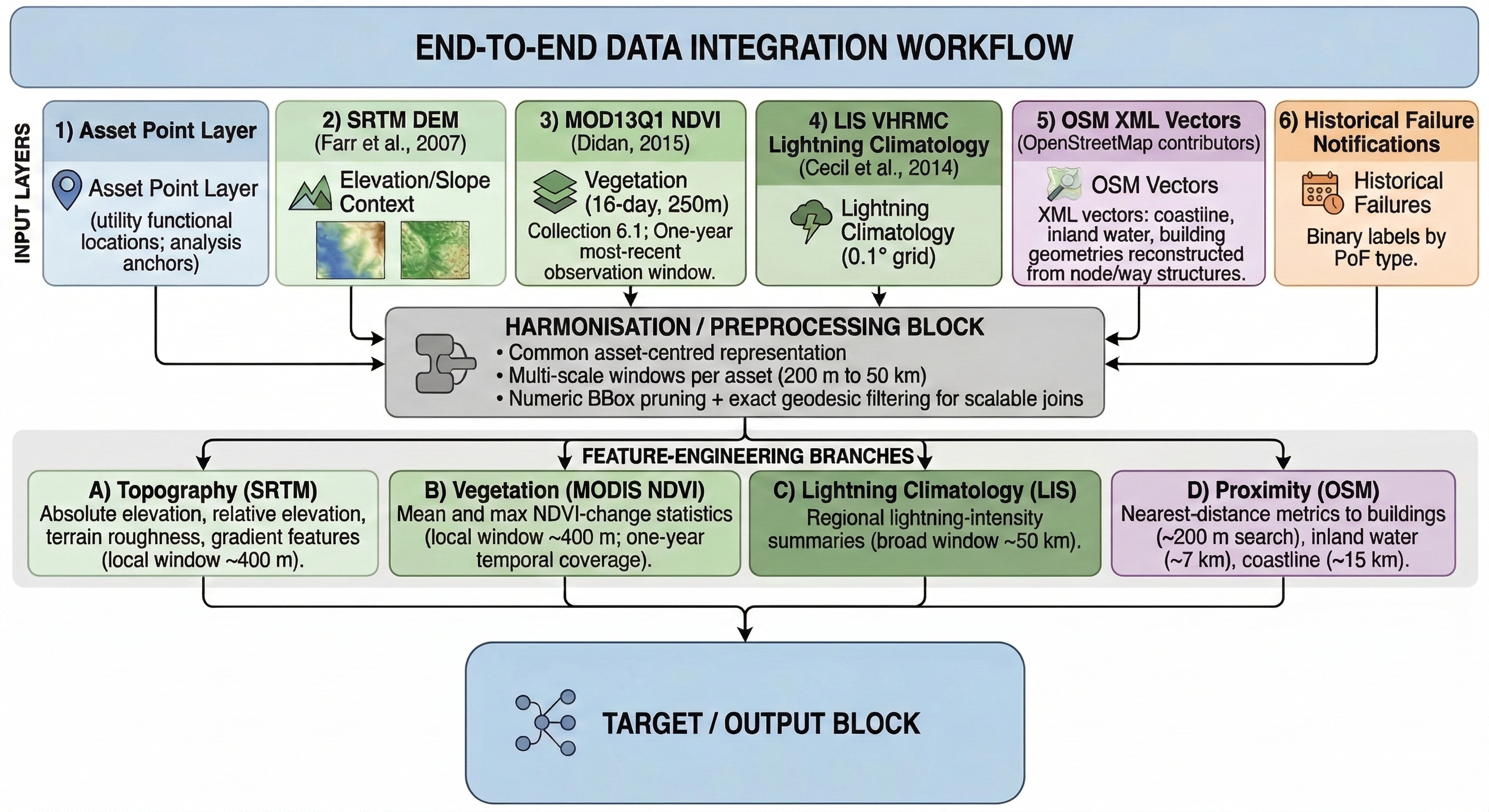}
\caption{Architecture of the data processing pipeline, illustrating the disparate handling of gridded environmental data and the reconstruction of vector geometries to assemble the final feature space.}
\label{fig:pipeline}
\end{figure}

The modelling framework integrates multi-source Earth-observation, volunteered geographic information, and utility operational data to characterise asset-level environmental exposure. Topographic context is represented using the NASA Shuttle Radar Topography Mission (SRTM) Global 3\,arc-second (V003) digital elevation model \citep{Farr2007SRTM}. Built-environment and land--water context are derived from OpenStreetMap (OSM) XML data \citep{OpenStreetMapContributors}, including coastline, inland-water, and building layers reconstructed into analysis-ready vector geometries. The conceptual data flow is illustrated if Fig~\ref{fig:pipeline}.

Vegetation conditions are represented using the MODIS/Terra Vegetation Indices product (MOD13Q1, Collection 6.1; 16-day composites, 250\,m, sinusoidal grid) \citep{Didan2021MOD13Q1C61}. From this product, we derive NDVI-based predictors as proxies for vegetation condition and spatial biomass dynamics, consistent with established ecological remote-sensing practice \citep{Pettorelli2005NDVI}. To align with industrial delivery constraints, NDVI features were computed from exactly one year of the most recent available observations. Lightning exposure is characterised using the LIS 0.1$^\circ$ Very High-Resolution Gridded Lightning Monthly Climatology (VHRMC; V1) \citep{Cecil2014LISClimatology}, which provides long-term spatial patterns of lightning occurrence. Supervised labels are obtained from the utility's historical failure-notification archive (binary failure/no-failure by asset and failure mode).

Taken together, these sources define the broader modelling feature space, spanning topographic attributes, proximity measures, vegetation and lightning indicators, prior-failure information, and asset/network-context variables. The full list of model features is summarised in Table~\ref{tab:feature_summary}, and the derivation of these variables is described below.

\subsection{Data preprocessing}

All datasets were harmonised into a common asset-centred geospatial representation prior to feature extraction. Asset coordinates were used as analysis anchors, and multi-scale regions of interest (200\,m to 50\,km) were constructed to capture both near-field effects (e.g., local relief, building proximity, short-range vegetation context) and broader mesoscale exposure (e.g., regional lightning climatology).

Raster data streams (SRTM, MODIS NDVI, and LIS VHRMC) were spatially queried within each analysis window and summarised through scale-appropriate descriptive statistics. Vector layers derived from OSM were spatially intersected with the same windows to compute proximity- and density-oriented predictors (e.g., distance to nearest water body, coastline, or building element). This harmonised processing ensures that predictors from heterogeneous sources are comparable at asset level and suitable for downstream machine-learning ingestion.

Failure-notification records were quality controlled, standardised, and linked to the corresponding asset identifiers to generate binary targets for each PoF type. The resulting analytical table combines engineered geospatial predictors, operational-history indicators, and asset/network-context variables in a single supervised-learning matrix. Table~\ref{tab:feature_summary} summarises the feature space at a conceptual level.

\begin{table}[H]
\centering
\caption{Full list of model features, descriptions, and spatial search constraints.}
\label{tab:feature_summary}
\resizebox{\textwidth}{!}{%
\begin{tabular}{@{}lllc@{}}
\toprule
\textbf{Category} & \textbf{Feature} & \textbf{Description} & \textbf{Search Radius} \\
\midrule
Topographic & Elevation Std. Dev. & Measure of local terrain roughness derived from the SRTM DEM. & 400\,m \\
Topographic & Relative Elevation & Difference between asset elevation and the local neighbourhood mean. & 400\,m \\
Topographic & Absolute Elevation & Absolute elevation at the asset location from the SRTM DEM. & 400\,m \\
Topographic & Elevation Gradient & Local terrain slope calculated from nearby DEM cells. & 400\,m \\
\midrule
Proximity & Distance to Coastline & Minimum geodesic distance to the nearest coastline. & 15\,km \\
Proximity & Distance to Waterbody & Minimum geodesic distance to the nearest inland waterbody. & 7\,km \\
Proximity & Distance to Building & Minimum geodesic distance to the nearest building. & 200\,m \\
\midrule
Environmental & Mean NDVI & Local mean NDVI as a proxy for vegetation condition. & 400\,m \\
Environmental & Max NDVI Change & Maximum positive NDVI change representing recent biomass growth. & 400\,m \\
Environmental & Lightning Intensity & Regional lightning-climatology intensity summary. & 50\,km \\
\midrule
Asset/Network & Aggregated Voltage & Aggregated asset voltage class. & Exact match \\
Asset/Network & SCONRR & Network-context classification (e.g., Rural, Metro, CBD). & Exact match \\
Asset/Network & Tree Overhang Flag & Indicator of observed tree overhang near the asset. & Exact match \\
\bottomrule
\end{tabular}%
}
\end{table}

\subsection{Data Processing and Spatial Optimisation}

The integration of high-resolution remote sensing grids with expansive vector databases (such as OSM) across a regional-scale utility network presents significant computational challenges. The primary objective of this data integration was to construct a comprehensive feature space for each utility asset, capturing its specific environmental exposure across key domains:
Key feature domains and representative engineered variables are summarised in Table~\ref{tab:feature_summary}.

To efficiently compute these features at scale, data preprocessing was executed end-to-end in the Snowflake cloud data platform. A highly optimised geospatial pipeline was developed in Snowflake to project, join, and aggregate these disparate spatial layers relative to the discrete point coordinates of each utility asset.

All geospatial transformations, feature engineering, intermediate table materialisation, and quality-control checks were performed within Snowflake. This unified cloud-native implementation ensured scalable distributed execution, consistent reproducibility across pipeline stages, and operational readiness for utility-scale deployment.

\subsubsection{Optimised Geospatial Pruning}

\begin{figure}[H]
\centering
\includegraphics[width=0.7\textwidth]{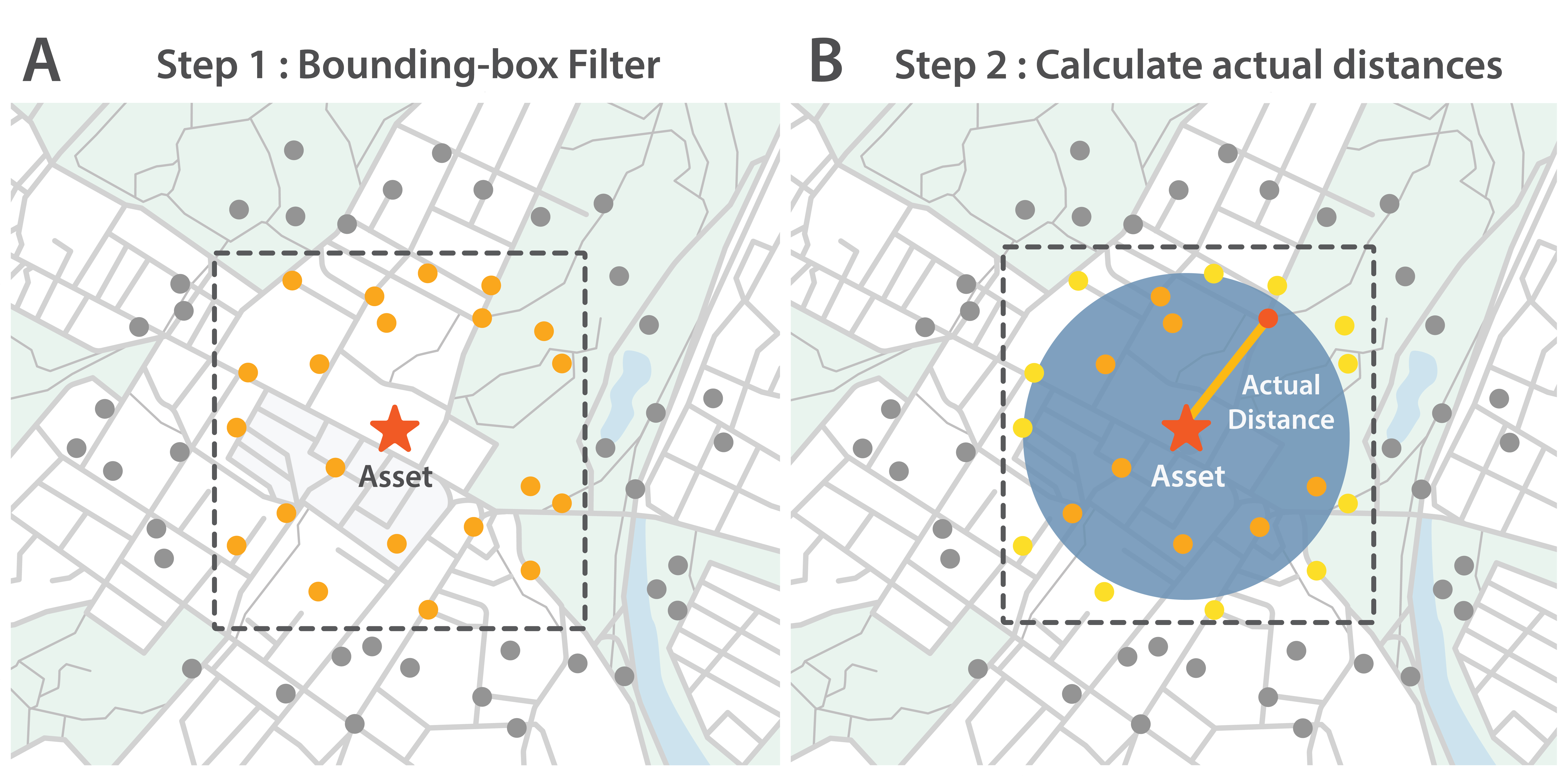}
\caption{Schematic representation of the two-step geospatial pruning algorithm. A dynamic latitudinal/longitudinal bounding box (Panel A) efficiently filters candidate points before exact geodesic distance calculations are applied (Panel B).}
\label{fig:spatial_pruning}
\end{figure}

A core methodological challenge was computing precise proximity metrics, such as the exact distance to the nearest coastline or vegetation stand, without executing computationally prohibitive Cartesian products across the entire regional dataset. To resolve this, we implemented a two-step geospatial pruning algorithm as shown in Fig~\ref{fig:spatial_pruning}.

First, a dynamic, numeric bounding box (BBox) was calculated for each asset based on the target search radius ($R$). To account for the Earth's curvature, the longitudinal delta ($\Delta \lambda$) was dynamically scaled by the latitude ($\phi$) of the asset point, effectively pre-filtering candidate geometries using simple numeric inequalities:
\begin{equation}
\Delta \lambda = \frac{R}{c \cdot \max(|\cos(\phi)|, \epsilon)}
\end{equation}
where $c \approx 111,320\,\mathrm{m}$ represents the nominal distance per degree of latitude, and $\epsilon$ is a safeguard constant to prevent division by zero near the poles.

Following this rapid numeric join, Snowflake geospatial SQL functions were applied only to the pruned candidate subsets. Specifically, \texttt{ST\_DWITHIN} was used as a spatial filter to retain only candidates within the target search distance of each asset, and \texttt{ST\_DISTANCE} was then used to calculate the exact final distance to the remaining candidates. This approach reduced computational overhead by orders of magnitude while preserving geometric accuracy.

\subsubsection{Vector Geometry Reconstruction}

Anthropogenic and hydrological boundaries from OpenStreetMap XML extracts required morphological reconstruction before distance calculations. Raw geographic nodes were parsed and linked to their corresponding spatial entities (e.g., natural coastline and building layers). Using sequential node indexing, discrete coordinate pairs were flattened and concatenated into continuous Well-Known Text (WKT) \texttt{LINESTRING} geometries. This vectorisation step enabled precise distance-to-edge measurements for localised assets, rather than less accurate centroid-based approximations.

\subsection{Geospatial Feature Engineering}

\begin{figure}[H]
\centering
\includegraphics[width=0.8\textwidth]{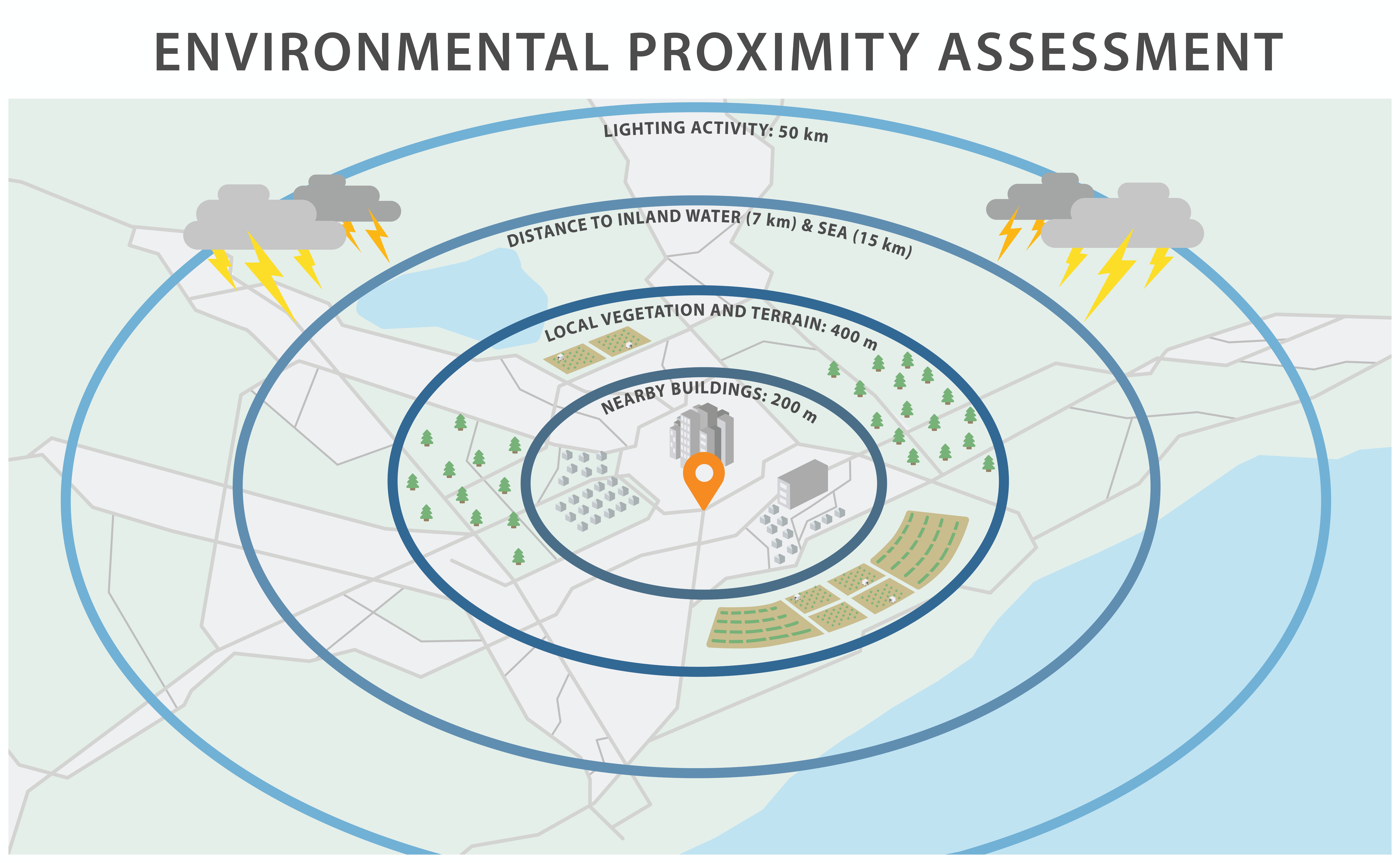}
\caption{Conceptual visualisation of multi-scale feature engineering. Concentric search radii define the spatial context for urban sheltering (200\,m), localised topographic anomalies (400\,m), and broader environmental exposure (up to 50\,km).}
\label{fig:feature_radii}
\end{figure}

Based on the spatial interactions between the infrastructure and the surrounding environment, a localised feature space was engineered for each functional location. The spatial search radii were tuned based on the morphological characteristics of the targeted variable. The high-level view is illustrated in Fig~\ref{fig:feature_radii}.

\paragraph{Topographic Features}
Terrain variability is a critical factor in localised weather exposure. We utilised a 400\,m spatial radius to extract absolute elevation profiles from the underlying DEM. From the localised grid, we derived standard deviation of elevation (to represent terrain roughness), relative elevation deviation (the asset's elevation minus the local 200\,m mean), and the nearest gradient slope.

\paragraph{Proximity and Anthropogenic Features}
Using the reconstructed OSM line geometries, exact spatial distances were calculated to contextualise the asset's immediate surroundings. We established varying search thresholds: 15\,km for coastlines (capturing maritime atmospheric influences), 7\,km for inland waterbodies, and 200\,m for building footprints (representing varying degrees of urban sheltering).

\paragraph{Environmental and Meteorological Features}
Vegetation density heavily influences the risk of secondary lightning impacts (e.g., localised fires or tree fall). The mean and maximum change of the Normalised Difference Vegetation Index (NDVI) were extracted within a 400\,m radius to quantify adjacent biomass vitality. These NDVI statistics were derived from a one-year, most-recent observation window. Finally, regional climatic exposure was captured by querying the maximum intensity of historical lightning strikes within a broad 50\,km radius, utilising the VHRMC climatology grid.

\subsubsection{Target Variable Formulation}

To construct the target variables for the subsequent binary classification model, historical maintenance and outage logs were integrated into the spatial feature matrix. Specifically, fault notifications from the utility's enterprise asset management system over a five-year period were extracted and analysed. Incidents definitively attributed to either lightning strikes or vegetation interference were isolated and matched to their respective functional locations.

These historical failure records were transformed into binary labels (i.e., the presence or absence of a vegetation or lightning-related fault at a given asset). It is critical to clarify the objective of this classification framework: rather than strictly predicting deterministic future failures at individual poles, these historical labels are utilised to train the model to assess the underlying probability of failure exposure given an asset's specific environmental, topographic, and spatial context. This probabilistic approach enables the identification of high-risk operational environments across the network, facilitating broad, targeted mitigation strategies over specific, localised predictions.

\subsection{Final Data Preparation and Imputation}

Prior to model ingestion, the assembled feature matrix required systematic treatment of missing values generated during the spatial joining processes (e.g., when no geometric feature was found within the defined search radius).

Continuous environmental variables, including terrain elevation gradients, local NDVI indices, and lightning climatology intensities, were subjected to mean imputation to represent average regional baselines in the absence of local anomalies. Conversely, unjoined spatial proximity measures required domain-specific constant imputation. Assets with no buildings or coastline within their respective search buffers were assigned values slightly beyond the maximum search distance (e.g., 17,000\,m for coastlines/buildings and 9,000\,m for waterbodies), effectively encoding them as "distant" without introducing algorithmic artefacts. Asset metadata (such as asset class and urban/rural categorisation) were explicitly cast as categorical data types to leverage native categorical handling algorithms.

\subsection{Machine Learning Architecture}

To model the non-linear interactions between the engineered geospatial features and asset failure probabilities, a Light Gradient Boosting Machine (LightGBM) architecture was selected \citep{Ke2017LightGBM}. LightGBM is highly effective for large-scale spatial datasets due to its gradient-based one-side sampling (GOSS) and exclusive feature bundling (EFB), alongside its native, optimised handling of discrete categorical variables without requiring highly sparse one-hot encodings.

While predicting multiple failure types could theoretically be consolidated into a single multi-target model, two distinct binary classification models were trained independently: one evaluating vegetation-related failure probability and the other evaluating lightning-related failure probability. This decoupled architecture was intentionally chosen to prioritise an industry-robust approach; isolating the target variables ensures that subsequent model analysis can be performed easily and diagnostic explainability, specifically using Shapley values, can be achieved with minimal operational overhead for domain experts. To avoid generating inferences on training data and ensure unbiased probability estimates, the modelling process utilised a 5-Fold Stratified Cross-Validation framework. This approach produced clean out-of-fold predictions, while the stratified sampling ensured that the minority class (fault occurrences) was proportionally represented across all training and validation splits, properly accounting for the highly imbalanced nature of utility failure datasets.

The models were regularised to prevent overfitting, utilising an intentionally shallow maximum tree depth of 4, a high minimum child sample threshold of 30, a learning rate of 0.03, and an ensemble size of 700 boosting iterations.

\subsubsection{Model Evaluation and Probabilistic Output}

Standard accuracy is often misleading in highly imbalanced spatial risk datasets. Therefore, model performance was evaluated using Out-Of-Fold (OOF) Area Under the Receiver Operating Characteristic Curve (ROC-AUC) and Out-Of-Fold Area Under the Precision--Recall Curve (PR-AUC). While ROC-AUC remains threshold-independent and informative for global class separability, PR-AUC is particularly suitable for rare-event settings because it directly emphasises positive-class precision and recall, making it more sensitive to minority-class performance under class imbalance \citep{Davis2006PRROC,Saito2015PRAUC}. The OOF methodology provides robust, unbiased performance estimates across the entire network geometry by aggregating validation predictions from all 5 independent folds.

To improve the transparency of the predictive spatial model, SHapley Additive exPlanations (SHAP) \citep{Lundberg2017SHAP} were calculated. SHAP values disaggregate the final probability outputs, allowing for an intuitive understanding of the relative importance and directional impact of each individual topographic, meteorological, and proximity feature on the asset's risk profile.

Finally, the raw probability outputs generated by the cross-validated models ($P_{OOF}$) represent the likelihood of a failure event occurring over the historical observation horizon ($H$). To make these insights operationally actionable for asset management, the probabilities were mathematically annualised ($P_{ann}$) under the standard constant-hazard assumption used in exponential reliability models, where survival over time $t$ is represented as $R(t)=e^{-\lambda t}$ \citep{NIST2003Reliability,MeekerEscobar1998Reliability}:

\begin{equation}
P_{ann} = 1 - (1 - P_{OOF})^{\frac{1}{H}}
\end{equation}

The resulting annualised probabilities provide a standardised risk metric, allowing utility planners to prioritise spatial clusters of assets facing elevated environmental exposure.

\section{Results}\label{sec:results}

This section presents the reported data diagnostics, model-validation results, and feature-effect analyses for the vegetation and lightning probability-of-failure models. It first provides an overview of the variable names used in the model-ready dataset, then summarises distributional diagnostics, cross-validated predictive performance, and model explainability outputs.

\subsection{Data}
For reference, the analytical variable names used in the model-ready dataset are listed in Table~\ref{tab:model_feature_names}. These names correspond directly to the higher-level feature groups summarised in Table~\ref{tab:feature_summary}.

\begin{table}[H]
\centering
\caption{Model feature names used in the analytical dataset.}
\label{tab:model_feature_names}
\resizebox{\textwidth}{!}{%
\begin{tabular}{@{}lll@{}}
\toprule
\textbf{Variable Name} & \textbf{Manuscript Feature} & \textbf{Category} \\
\midrule
\texttt{FEATURE\_1\_ELEVATION\_STDEV\_M} & Elevation Std. Dev. & Topographic \\
\texttt{FEATURE\_2\_ELEVATION\_DEVIATION\_M} & Relative Elevation & Topographic \\
\texttt{FEATURE\_3\_NEAREST\_ELEVATION\_M} & Absolute Elevation & Topographic \\
\texttt{FEATURE\_4\_NEAREST\_GRADIENT} & Elevation Gradient & Topographic \\
\texttt{FEATURE\_5\_DISTANCE\_TO\_SEA\_M} & Distance to Coastline & Proximity \\
\texttt{FEATURE\_6\_DISTANCE\_TO\_WATER\_M} & Distance to Waterbody & Proximity \\
\texttt{FEATURE\_7\_DISTANCE\_TO\_BUILDING\_M} & Distance to Building & Proximity \\
\texttt{FEATURE\_8\_NEAREST\_MEAN\_NDVI} & Mean NDVI & Environmental \\
\texttt{FEATURE\_8\_NEAREST\_MAX\_CHANGE\_NDVI} & Max NDVI Change & Environmental \\
\texttt{FEATURE\_9\_NEAREST\_LIGHTNING\_INTENSITY} & Lightning Intensity & Environmental \\
\texttt{AGGREGATED\_VOLTAGE\_KV} & Operational Voltage & Asset/Network \\
\texttt{SCONRR} & SCONRR region & Asset/Network \\
\texttt{IS\_TREE\_OVERHANG} & Tree Overhang Flag & Asset/Network \\
\bottomrule
\end{tabular}%
}
\end{table}

Due to confidentiality constraints, we limit the reported diagnostics to non-sensitive distributions only. In this section, we provide a concise narrative summary of key encoded variables and qualitative distribution signals used for data-quality checks and model-readiness assessment. Additionally, we share the feature distribution panels in Appendix~\ref{app:features}.

SCONRR region is encoded as \{0: Rural, 1: Metro, 2: CBD\}. The full vegetation and lightning distribution panels are provided in Figures~\ref{fig:veg_feature_distributions_app} and \ref{fig:lightning_feature_distributions_app}, respectively.

The data diagnostics indicate that SCONRR captures distinct network contexts (Rural, Metro, and CBD). The tree-overhang indicator, SCONRR, and aggregated voltage class display clear class-separation patterns in the distributions. These contrasts are consistent with the feature-effect interpretation in Section~\ref{sec:FeatEffandExplain}. In addition, the full distribution panels indicate class imbalance for both hazard targets.

\subsection{Model performance and validation}
\begin{figure}[H]
\centering
\includegraphics[width=0.85\textwidth]{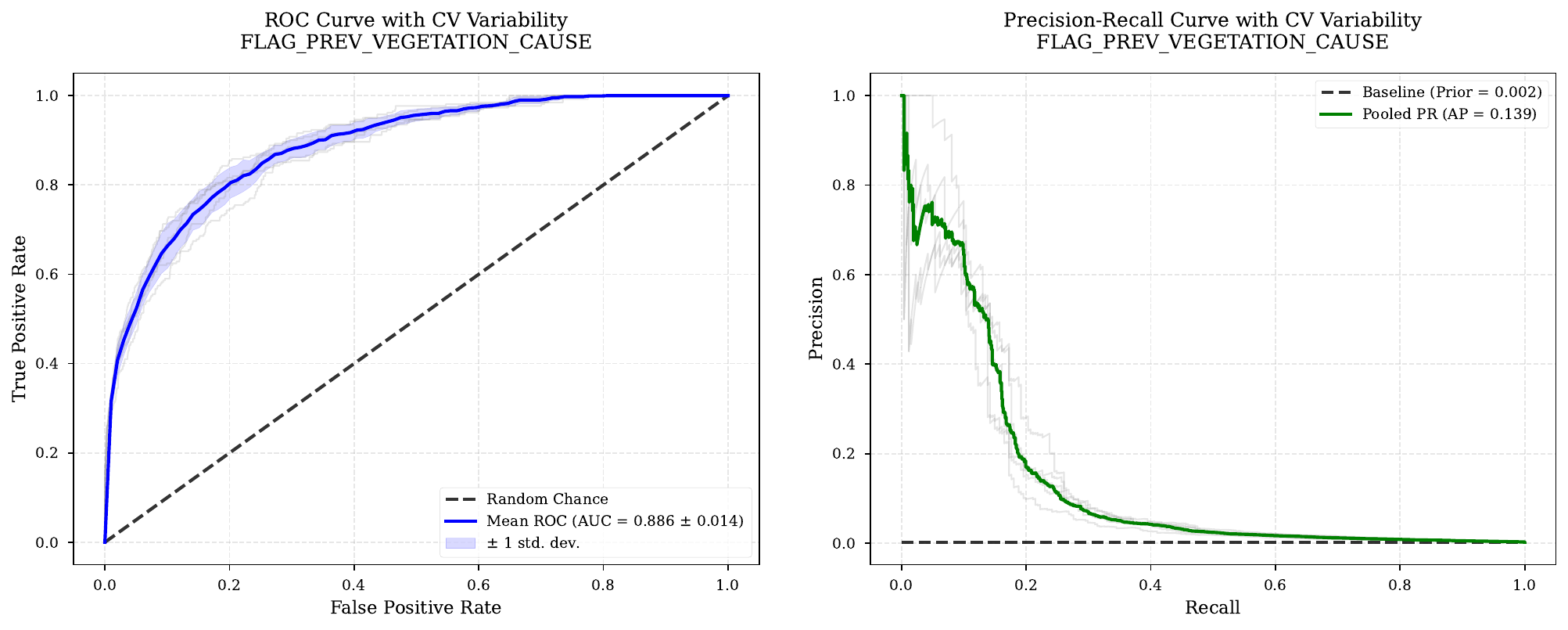}
\caption{Cross-validated ROC and Precision--Recall curves for the vegetation-related failure model.}
\label{fig:veg_roc_pr}
\end{figure}

\begin{figure}[H]
\centering
\includegraphics[width=0.85\textwidth]{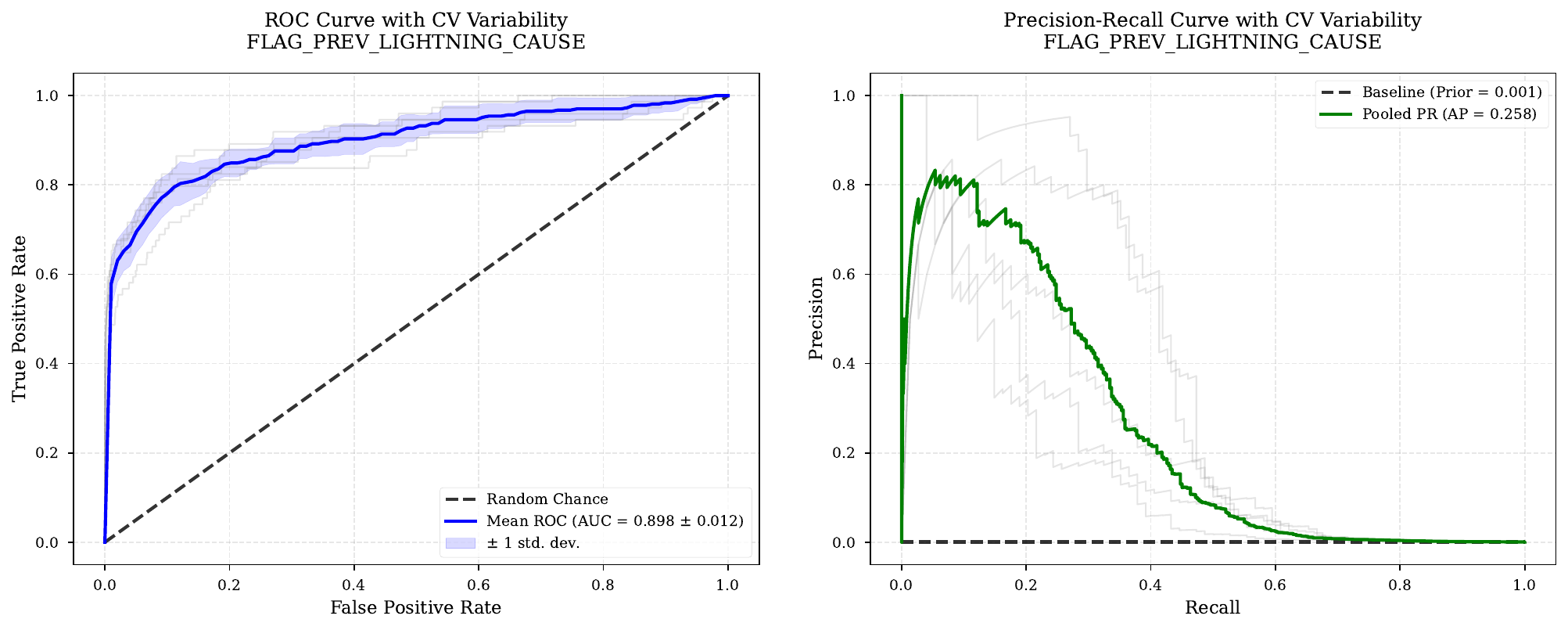}
\caption{Cross-validated ROC and Precision--Recall curves for the lightning-related failure model.}
\label{fig:lightning_roc_pr}
\end{figure}

Cross-validated discrimination performance is shown separately for the vegetation and lightning models in Figures~\ref{fig:veg_roc_pr} and \ref{fig:lightning_roc_pr}. 
The ROC and Precision--Recall curves indicate stable fold-level behaviour and support the suitability of the framework for imbalanced utility-failure prediction tasks. Notably, fold-level stability is lower for the lightning model, which is most visible in the PR-AUC curves.  
We highlight that the observed model performance is conditioned by the geolocation-unaware training/test data split; this is further discussed in the Limitations subsection of Section~\ref{sec:Discussion}.

\subsection{Feature effects and model explainability}\label{sec:FeatEffandExplain}
\begin{figure}[H]
\centering
\includegraphics[width=0.95\textwidth]{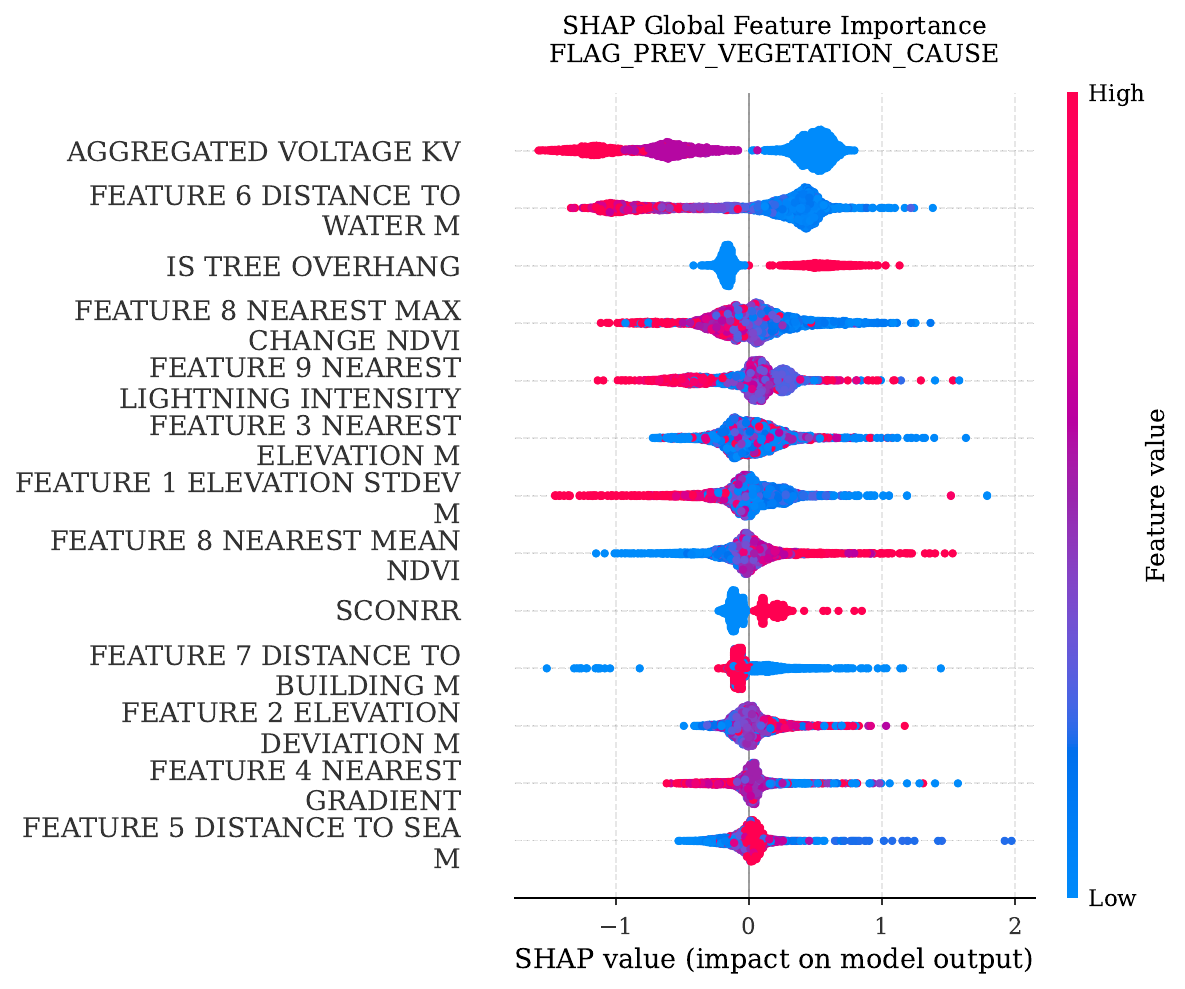}
\caption{Global SHAP summary for the vegetation-related failure model.}
\label{fig:veg_shap_global}
\end{figure}

\begin{figure}[H]
\centering
\includegraphics[width=0.95\textwidth]{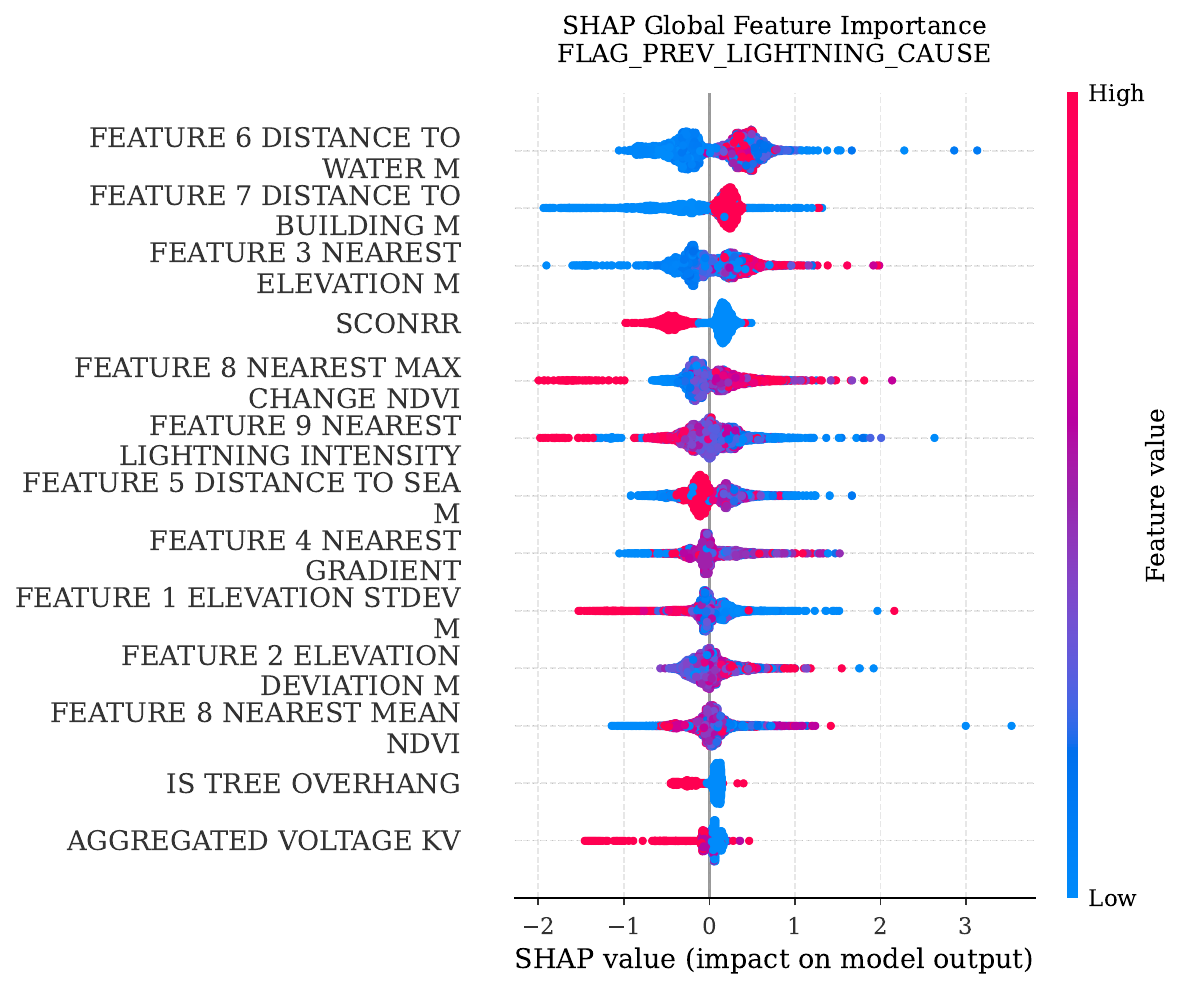}
\caption{Global SHAP summary for the lightning-related failure model.}
\label{fig:lightning_shap_global}
\end{figure}

SHAP global summaries for both hazard-specific models are shown in Figures~\ref{fig:veg_shap_global} and \ref{fig:lightning_shap_global}. Across both models, topographic context, vegetation-condition indicators, lightning-climatology variables, and spatial proximity measures contribute strongly to risk stratification, while the relative ordering of feature importance differs by failure mechanism.

For the vegetation failure model, the three most influential features are asset voltage, distance to water, and the tree-overhang indicator. Additional separation is observed for SCONRR category and local elevation standard deviation. 

The vegetation model indicates that lower aggregated voltage classes are associated with higher predicted vegetation-failure risk. Shorter distance to water and the presence of tree overhang are also associated with elevated risk, consistent with their high SHAP contribution. In addition, higher mean NDVI values correspond to increased predicted risk, and larger SCONRR values (Metro and CBD categories) are similarly associated with higher model outputs. 

For the lightning failure model, the three most influential features are distance to water, distance to the nearest building, and local asset elevation relative to surrounding terrain. Additional separation is observed for SCONRR category, the tree-overhang indicator, and asset voltage, with the latter two showing smaller effects on model output.

Inspection of the lightning-model SHAP patterns indicates the following:
\begin{itemize}
\item Shorter distance to buildings is associated with lower predicted lightning-failure risk.
\item Rural SCONRR region class is associated with higher risk relative to Metro/CBD classes.
\item Higher local absolute elevation at the asset location is associated with increased predicted risk.
\item Higher local elevation variability (terrain ruggedness) is associated with reduced risk, indicating elevated risk for assets in flatter surrounding terrain.
\item The presence of tree overhang and higher-voltage equipment is associated with modest risk reduction in this model.
\end{itemize}

\section{Discussion}\label{sec:Discussion}

The developed framework demonstrates that multi-source geospatial feature engineering and gradient-boosted probabilistic modelling can support more targeted asset risk management under lightning and vegetation exposure. By integrating terrain, vegetation, hydrological, and built-environment predictors at multiple spatial scales, the approach provides interpretable risk estimates that can complement conventional rule-based maintenance planning.

\subsection{Analysis of Findings}
The SHAP-based feature contributions indicate that the models capture physically and operationally plausible exposure mechanisms rather than relying on purely statistical artefacts. For vegetation-related failures, the dominant predictors (asset voltage class, distance to water, and tree-overhang flag) align with known field conditions. Shorter distance to water is a proxy for more persistent vegetation growth potential and higher branch-fall susceptibility in some species (e.g., eucalyptus/gum trees under high-moisture growth conditions), while direct overhang is an immediate mechanism for contact-related events. The positive contribution of higher NDVI similarly aligns with increased local biomass availability. The SCONRR effect suggests that network context (Rural/Metro/CBD) modulates baseline exposure through differences in asset surroundings and operating environments. In particular, Metro and CBD areas are subject to stricter regulations that can limit pre-emptive vegetation management for low-voltage lines, which may contribute to higher residual vegetation-related risk.

For lightning-related failures, the SHAP pattern indicates two interacting drivers: broader regional exposure and immediate asset siting characteristics. Higher local absolute elevation contributes positively to risk, consistent with greater strike susceptibility of more exposed assets. Greater terrain ruggedness around an asset is associated with lower modelled risk. Distance-to-building effects may reflect shielding or correlated infrastructure context in denser built environments. The SCONRR region contribution appears strongly indicative of whether assets benefit from city-infrastructure lightning-protection conditions, making it a key modifier of lightning risk.

Taken together, these patterns support the practical interpretability of the framework: the dominant predictors are not only statistically influential but also actionable for utility decision-making, because they map to interventions such as vegetation management, local inspection prioritisation, and location-aware hardening strategies. 

\subsection{Limitations}
This study has several limitations that should be considered when interpreting the reported performance and operational implications.

First, the modelling strategy is deliberately scoped rather than benchmark-driven. We use a modular architecture with separate binary models for each PoF type, and we do not perform formal benchmarking against alternative model families (e.g., random forests, XGBoost, neural networks). This choice prioritises a scalable, interpretable, and operationally maintainable framework under utility constraints; however, it means that the present study does not claim that the selected model class is universally optimal.

Second, the validation design does not enforce explicit geospatial or temporal train--test separation. Instead, we use stratified cross-validation to obtain statistically efficient estimates under severe class imbalance. As a result, reported metrics may be optimistic for strict deployment scenarios involving transfer to unseen regions or forecasting into future periods, where spatial autocorrelation and temporal dependence are typically weaker between training and test distributions. 
It should be emphasised that rigorous spatial-block and temporal-holdout evaluation requires long, consistently structured historical records that are not always available in utility settings. This data-availability constraint is therefore a practical limitation of both this case study and the broader domain.

Third, NDVI temporal coverage is limited to one year of recent observations due to delivery constraints. Although this window supports operational implementation, it may under-represent inter-annual vegetation variability, lagged ecological responses, and rare climatic conditions; consequently, vegetation-related PoF estimates may be less sensitive to longer-term environmental dynamics than models trained on multi-year vegetation histories.

Overall, these design choices remain consistent with the study objective: to establish and demonstrate a scalable, extensible, asset-level PoF framework that is operationally useful in a real utility context, rather than to provide definitive estimates of out-of-region transferability or long-horizon forecasting skill.

\subsection{Future Work}
A key next step is transitioning from retrospective risk estimation to near-real-time risk monitoring. The MODIS/Terra Vegetation Indices product (MOD13Q1), used here for NDVI-derived predictors, is updated at 16-day intervals, enabling regular refresh of vegetation-condition signals. In parallel, integrating temporally resolved lightning observations, rather than long-term climatology alone, would allow explicit modelling of short- to medium-term lightning activity trends. Together, these updates create a pathway to a near-real-time risk monitoring platform that can re-prioritise assets as environmental conditions evolve.

A second high-value extension is vision-driven feature generation from operational imagery. Utility inspection photos can be analysed with computer-vision pipelines to extract structured indicators such as conductor--vegetation clearance, branch overhang severity, pole or hardware condition, and local obstruction context. These image-derived variables could be fused with the current geospatial predictors to improve both near-field exposure characterisation and operational explainability.

Relatedly, street-level panoramic imagery (e.g., Google Street View or equivalent platforms) could support scalable external feature extraction to enhance direct inspection imagery. Such data can provide repeatable corridor-level context for vegetation encroachment, built-environment shielding, and asset-visibility proxies at large scale. Future work can evaluate multimodal modelling that combines satellite, tabular geospatial, LiDAR, and street view inputs under a common asset-level risk framework.
This direction is especially relevant under climate change, where shifts in extreme-weather regimes, fuel conditions, and storm behaviour may alter baseline failure risk over time and require adaptive, continuously updated decision support.

An implication of the condition-based design is its potential transferability across geographies. Because the model is structured around observable asset conditions and environmental exposure variables, rather than only location-specific identifiers, patterns that are rare or out-of-distribution in one service territory may already be represented in the historical data of another. This is particularly important under climate change: conditions that have not yet been common in one region may begin to materialise over time, while analogous combinations of heat, vegetation condition, terrain, water proximity, and lightning exposure may already exist elsewhere. Although formal geographic transfer testing is outside the scope of this study, the modular feature design highlights a pathway toward cross-region learning, where utilities can use broader historical experience to anticipate emerging local risks without reworking the core architecture.

\section{Conclusion}\label{sec:conclusion}
This study demonstrates that a modular, geospatially harmonised machine-learning framework can deliver operationally meaningful asset-level PoF estimation for both vegetation-related and lightning-related failure modes. By integrating multi-source Earth-observation data, built-environment context, and utility failure records, the approach captures hazard-specific exposure patterns while maintaining a shared, scalable modelling pipeline.

The results indicate that the dominant predictors are interpretable and actionable for field deployment, supporting targeted interventions such as vegetation management, inspection prioritisation, and location-aware hardening. At the same time, the reported performance should be interpreted within study scope, including the absence of strict spatial/temporal holdouts and formal cross-model benchmarking.

Overall, the framework provides a practical foundation for risk-informed utility asset management and can be extended toward near-real-time decision support through temporally refreshed environmental inputs and vision-driven features from inspection and street-level imagery.

\section*{Acknowledgements}
The authors thank Ali Walsh for her support throughout the project. 
The authors also thank Ashleigh Lamb for her design work on the figures.

\bibliographystyle{elsarticle-harv} 
\bibliography{literature.bib}

\FloatBarrier

\appendix
\begin{appendices}

\section{Glossary of acronyms}\label{app:glossary}
\begin{description}
\item[CBD] Central Business District.
\item[DEM] Digital Elevation Model.
\item[LIS] Lightning Imaging Sensor.
\item[LightGBM] Light Gradient Boosting Machine.
\item[MOD13Q1] MODIS/Terra Vegetation Indices product identifier.
\item[MODIS] Moderate Resolution Imaging Spectroradiometer.
\item[NASA] National Aeronautics and Space Administration.
\item[NDVI] Normalised Difference Vegetation Index.
\item[OOF] Out-of-fold.
\item[OSM] OpenStreetMap.
\item[PoF] Probability of failure.
\item[PR] Precision--Recall.
\item[ROC] Receiver Operating Characteristic.
\item[SAPN] SA Power Networks.
\item[SCONRR] Steering Committee on National Regulatory Reporting. Utility network-context region classification variable used in the asset data (e.g., Rural, Metro, CBD).
\item[SHAP] SHapley Additive exPlanations.
\item[SQL] Structured Query Language.
\item[SRTM] Shuttle Radar Topography Mission.
\item[VHRMC] Very High-Resolution Gridded Lightning Monthly Climatology.
\item[WKT] Well-Known Text.
\item[XML] Extensible Markup Language.
\end{description}

\section{Feature set}\label{app:features}
\begin{figure}[H]
\centering
\includegraphics[width=0.95\textwidth]{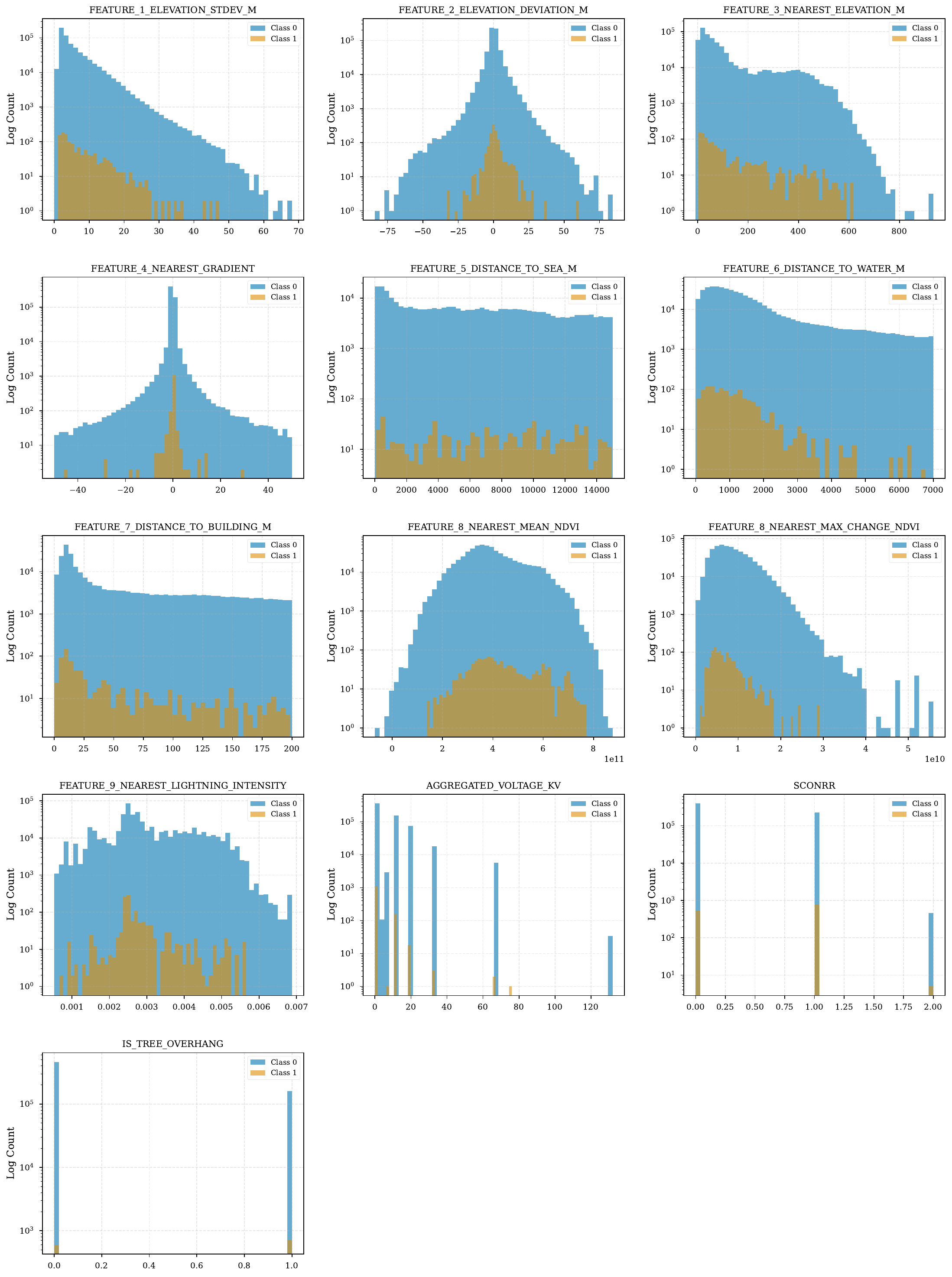}
\caption{Feature distributions for the vegetation-related failure model input space.}
\label{fig:veg_feature_distributions_app}
\end{figure}

\begin{figure}[H]
\centering
\includegraphics[width=0.95\textwidth]{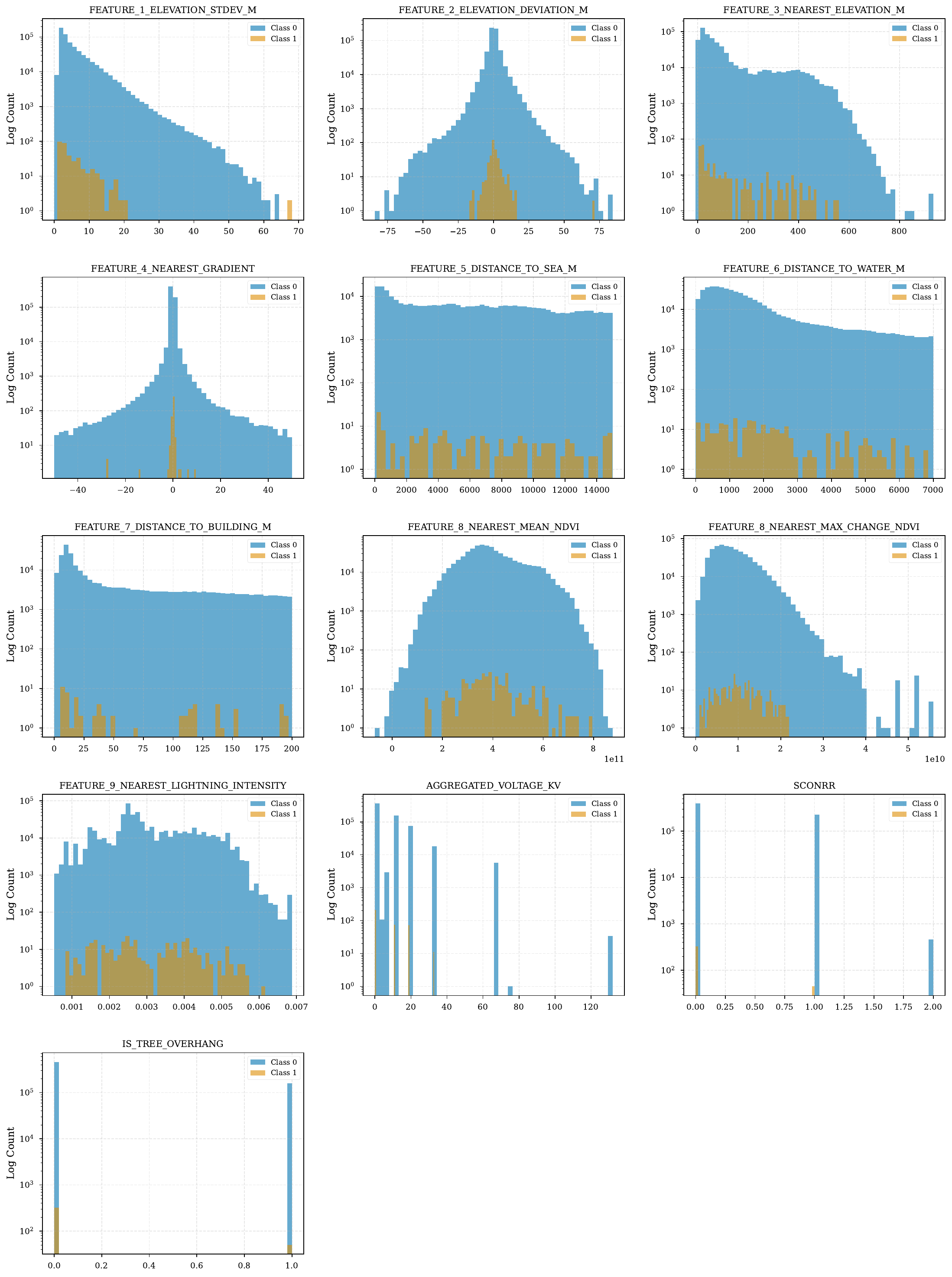}
\caption{Feature distributions for the lightning-related failure model input space.}
\label{fig:lightning_feature_distributions_app}
\end{figure}

Figures~\ref{fig:veg_feature_distributions_app} and \ref{fig:lightning_feature_distributions_app} provide the full feature-distribution diagnostics for the vegetation and lightning failure models, respectively.

\end{appendices}

\end{document}